\documentclass[10pt, logo, twocolumn, copyright]{nvidiatechreport}

\usepackage{mdframed}
\usepackage[utf8]{inputenc} 
\usepackage[T1]{fontenc}    

\usepackage{amsfonts}       
\usepackage{nicefrac}       
\usepackage{microtype}      
\usepackage[dvipsnames]{xcolor}         
\usepackage{multirow}
\usepackage{multicol}
\usepackage{graphicx}
\usepackage[numbers]{natbib}
\usepackage{tabto}
\usepackage{xspace}
\usepackage{amsmath}
\usepackage{adjustbox}
\usepackage{enumitem}
\usepackage{wrapfig}
\usepackage{dblfloatfix}
\usepackage{makecell}
\usepackage{subcaption}

\usepackage{footmisc}
\usepackage{tcolorbox}
\usepackage{enumitem}
\usepackage{float}

\usepackage[ruled,vlined,linesnumbered]{algorithm2e}

\usepackage{natbib}

\usepackage{amsmath,amsfonts,bm}

\def\eqref#1{equation~\ref{#1}}

\def\1{\bm{1}}

\DeclareMathAlphabet{\mathsfit}{\encodingdefault}{\sfdefault}{m}{sl}
\SetMathAlphabet{\mathsfit}{bold}{\encodingdefault}{\sfdefault}{bx}{n}

\usepackage{hyperref}
\usepackage{url}
\usepackage{booktabs}
\usepackage{multirow}
\usepackage{enumitem}
\usepackage{adjustbox}
\usepackage{tabularx}
\usepackage{arydshln}
\usepackage{wrapfig}

\usepackage{multirow}
\usepackage{colortbl}
\usepackage{amsmath}
\usepackage{makecell}
\usepackage{hhline}
\usepackage{array}
\usepackage{diagbox}
\usepackage{graphicx}
\usepackage{adjustbox}

\usepackage{colortbl} 
\usepackage[table]{xcolor} 

\usepackage[flushleft]{threeparttable} 
\usepackage{multirow}

\usepackage{wrapfig}

\definecolor{Note_color}{rgb}{0.0, 0.0, 0.0}

\definecolor{lightyellow}{RGB}{255, 255, 204}
\definecolor{lightred}{RGB}{255, 204, 204}
\definecolor{lightgreen}{RGB}{204, 255, 204}

\usepackage{xcolor}
\definecolor{maskorange}{HTML}{ffe6cc}
\definecolor{maskblue}{HTML}{d4e1f5}
\definecolor{maskred}{HTML}{ffcccc}
\definecolor{maskgreen}{HTML}{d5e7d4}
\definecolor{diff_green}{HTML}{7cb342}
\definecolor{nldgreen}{HTML}{E8F5E9}

\usepackage{pifont}
\newcommand\mynuma[1]{\ifcase#1 \or \ding{172}\or \ding{173}\or
  \ding{174}\or \ding{175}\or \ding{176}\or \ding{177}
  \or \ding{178}\or \ding{179}\or \ding{180}\or \ding{181}\else *\fi\relax}

\newcommand\mynumb[1]{\ifcase#1 \or \ding{182}\or \ding{183}\or
  \ding{184}\or \ding{185}\or \ding{186}\or \ding{187}
  \or \ding{188}\or \ding{189}\or \ding{190}\or \ding{191}\else *\fi\relax}

\usepackage{booktabs}
\usepackage{amsmath,amssymb}
\usepackage{multirow}
\usepackage{xcolor}
\usepackage{placeins}   
\usepackage[capitalise]{cleveref}

\graphicspath{{figures/}}

\usepackage{caption}

\NewDocumentCommand{\yonggan}
{ mO{} }{\textcolor{red}{\textsuperscript{\textit{yonggan}}\textsf{{\small[#1]}}}}

\NewDocumentCommand{\pavlo}
{ mO{} }{\textcolor{red}{\textsuperscript{\textit{pavlo}}\textsf{{\small[#1]}}}}

\NewDocumentCommand{\siyan}
{ mO{} }{\textcolor{green!50!black}{\textsuperscript{\textit{siyan}}\textsf{{\small[#1]}}}}

\newcommand{\method}{Semi-OPD\xspace}

\title{When Do We Need On-Policy Distillation? Distilling on Offline Student Rollouts Is Often Better}

\author{
Siyan Zhao$^{*\dagger1}$, Yonggan Fu$^{*}$, Jindong Jiang, Shih-Yang Liu, Song Bian, Byung-Kwan Lee, Sharath Turuvekere Sreenivas, Wenliang Dai, Hanrong Ye, Aditya Grover$^{1}$, Pavlo Molchanov
}

\correspondingauthor{X}

\begin{abstract}
On-policy distillation (OPD) has become increasingly popular for transferring teacher capabilities to student models. In this work, we ask a critical research question: Is on-policy sampling always beneficial for distilling arbitrary teacher--student pairs? We show that a simple alternative, \method, which distills from offline rollouts generated by the initial student, can often outperform OPD in both accuracy and training efficiency. Across 17 teacher--student pairs ranging from 1.5B to 235B parameters, \method outperforms OPD in 14 cases, with up to +13.6\% accuracy and 11.4$\times$ training speedup. 
We further find that the choice between OPD and \method depends on the alignment between the initial teacher and student, quantified by an output-token overlap ratio: OPD is beneficial only when the two are highly aligned with high overlap ratios.
Our deeper investigation suggests that effective distillation requires on-policyness w.r.t. both the student and the teacher. For misaligned pairs, student rollouts can become increasingly off-policy w.r.t. the teacher as context length grows, weakening the distillation signal. In contrast, \method is often more stable, as it distills on shorter contexts while covering full trajectories and exposing the student to more teacher-preferred tokens.
Beyond proposing \method as an efficient alternative, our work motivates the community to rethink when to use OPD and to study stronger OPD variants with meaningful teacher--student pairs.

\end{abstract}

\newcommand{\teaserfigure}{%
  \begingroup
  \captionsetup{type=figure}%
  \centering
  \includegraphics[width=0.95\linewidth]{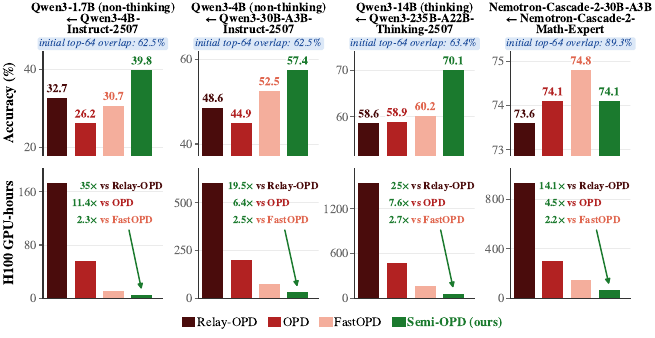}
  \vspace{-0.5em}
\caption{
\textbf{Accuracy \& training cost of different distillation methods.} Our \method achieves much improved/comparable accuracy for teacher--student pairs with low/high overlap, respectively, while reducing training time by up to $11.4\times$ over OPD. 
Accuracy is averaged over AIME~24/25/26 and HMMT~Feb~26.
}
\label{fig:speedup}
  \endgroup
}

\makeatletter
\renewcommand{\maketitle}{%
  \twocolumn[%
    \begin{adjustwidth}{0pt}{24pt}%
      \begin{flushleft}%
        {\raggedright \titlefont \@title\par}%
        \vskip11pt%
        {\raggedright \@author\par}%
        \vskip20pt%
      \end{flushleft}%
    \end{adjustwidth}%
    \abscontent
    \vskip16pt
    \teaserfigure
    \vskip20pt%
  ]%
  \thispagestyle{firststyle}%
}
\makeatother

\begin{document}
\maketitle

\begin{figure*}[t]
  \centering
  \includegraphics[width=0.95\linewidth]{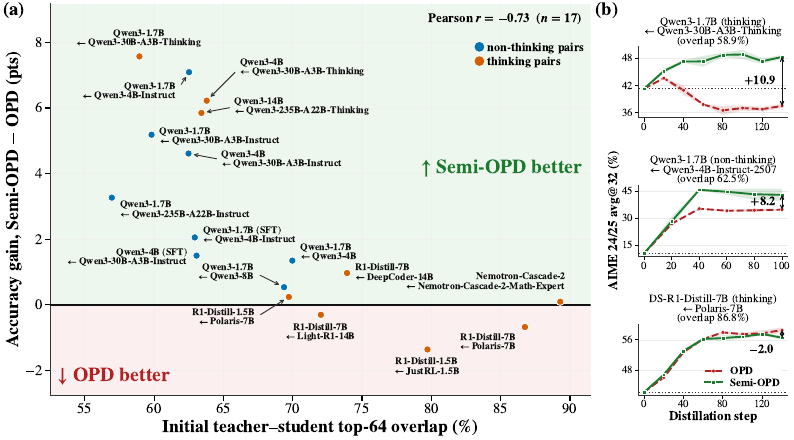}
  \vspace{-0.5em}
  \caption{
\textbf{Semi-OPD outperforms standard OPD when the initial teacher--student overlap is low.}
\textbf{(a)} Accuracy gain vs.\ initial top-64 overlap across 17 pairs ranging from 1.5B to 235B,
spanning thinking and non-thinking modes and multiple model families, with a Pearson
correlation of $r=-0.73$. Accuracy gain is the area between the \method and OPD accuracy curves
divided by their shared training horizon, averaged over AIME~24/25/26 and HMMT~Feb~26.
  \textbf{(b)}~Training curves for three pairs of increasing overlap (AIME~24/25 avg@32,
  mean of 3 seeds). At low overlap, \method outperforms OPD which can even degrade;
  at high overlap, \method can slightly lag behind.}
  \label{fig:overlap}
  \vspace{-1em}
\end{figure*}

\section{Introduction}
\label{sec:intro}

On-policy distillation (OPD) has emerged as an effective paradigm for transferring the capabilities of a teacher language model (LM) to a student LM~\citep{agarwal2024policy, gu2024minillm, lu2025onpolicydistillation, xiao2026mimov2flashtechnicalreport, xu2026deepseek}. It mitigates the training--inference distribution mismatch, i.e., exposure bias, in traditional knowledge distillation by providing dense token-level signals from the teacher LM on student rollouts obtained via on-policy sampling. 

Despite the promise, recent works reveal several artifacts of OPD:~\citep{luo2026demystifying, yang2026eos, fu2026revisiting, yu2026mismatch} find that on-policy rollouts tend to grow longer and fall into self-correction loops, exhausting the generation budget without emitting an EOS token;~\citep{zhang2026fast,ziheng2026less, liu2026prefix} find that supervision on later tokens in student rollouts is often unhelpful and that truncating to a short prefix can suffice; and~\citep{xu2026pass,jiang2026trajectory} similarly finds that once the student commits to an incorrect reasoning direction, subsequent misdirected continuations elicit unreliable supervision and waste compute. These artifacts lead to our central question:

\begin{center}
\vspace{-0.5em}
\emph{Is on-policy sampling always beneficial for distilling arbitrary teacher–student pairs?}
\vspace{-0.5em}
\end{center}

To answer this, we propose a simple and efficient alternative called semi-on-policy distillation (\method), where rollouts are sampled once from the \emph{initial} student before any training and reused throughout distillation, thus making both rollout and teacher logit generation offline. Although the source of rollouts is fixed, the objective is identical to OPD's, and the student's log-probabilities, and hence the per-token advantage, are recomputed using the current parameters at every step.
We evaluate across 17 teacher--student pairs spanning 1.5B to 235B parameters and three model families in both thinking and non-thinking modes, covering the most commonly used student--teacher pairs in the OPD literature. We find that \method outperforms OPD on 14 of them, with up to +13.6\% accuracy and 11.4$\times$ training speedup as shown in Fig.~\ref{fig:speedup}, and achieves comparable performance (within a 0.1\%--2.6\% gap)  on the remaining three.
This phenomenon motivates our further investigation into how to choose between \method and OPD and why \method is often better.

\noindent \textbf{How to choose between \method and OPD.}
We find that the gain of \method over OPD is strongly correlated with the alignment between the initial teacher and student, which can be quantified by the teacher--student top-$k$ overlap ratio over output tokens, as shown in Fig.~\ref{fig:overlap}a.
The accuracy evolution curves in Fig.~\ref{fig:overlap}b further show that (1) when this overlap ratio is low, continually regenerating rollouts from the evolving student can shift training toward increasingly teacher-misaligned trajectories, making strict on-policyness counterproductive; (2) OPD is only beneficial when the teacher and student are aligned with a high overlap ratio.

\noindent \textbf{Why OPD often underperforms \method.} Our empirical study suggests that effective distillation requires ``on-policyness'' w.r.t. both the student and the teacher, where the former mitigates exposure bias and the latter determines the effectiveness of the distillation signal. OPD improves the former but may reduce the latter.
Specifically, for misaligned pairs, student rollouts from OPD can become increasingly off-policy w.r.t. the teacher as context length grows. This cannot be addressed by simple truncation, as we empirically show the importance of covering supervision signals from different reasoning stages.
In contrast, \method on initial student rollouts usually distills over shorter contexts while covering full trajectories, making it more stable. In other words, \method trades some student on-policyness for teacher on-policyness, which is also echoed by our observation that \method exposes the student to more tokens preferred by the teacher.

We aim to inspire the community to rethink when to use OPD and to study stronger OPD variants under meaningful settings. The key takeaways of our work are summarized below:

\begin{itemize}[leftmargin=1.2em]
\vspace{-1em}
\item OPD is only useful for aligned teacher--student pairs with high overlap in output tokens; for misaligned pairs, \method is a better alternative in both accuracy and training efficiency.

\item Effective distillation requires both the on-policyness of training rollout w.r.t. the student and closeness to the teacher; OPD prioritizes the former but may sacrifice the latter.

\item Given the target domain and initial teacher--student pair, the choice between OPD and \method should be guided by the measured overlap ratio, rather than blindly choosing OPD.

\item For future research on OPD, teacher--student pairs should be carefully selected to ensure a meaningful experimental setting.

\end{itemize}

\vspace{-0.5em}
\section{An Efficient OPD Alternative: Semi-On-Policy Distillation}
\vspace{-0.5em}

\label{sec:method}

\subsection{Preliminaries of OPD}

Formally, given a prompt $x \sim \mathcal{D}$, a student LLM $\pi_\theta$ generates a
response $y=(y_1,\dots,y_T)$ autoregressively, visiting prefix states $s_t=(x,y_{<t})$.
The underlying sequence-level objective is the reverse
$\mathrm{KL}(\pi_\theta(\cdot\mid x)\,\|\,\pi_T(\cdot\mid x))$ with respect to a fixed teacher
$\pi_T$. In practice, a common approximation uses a local token-level objective that
evaluates the teacher--student log-probability ratio
only on tokens sampled by the student, avoiding a full-vocabulary comparison at
every position~\citep{lu2025onpolicydistillation, team2026kimi}.
This yields the sampled-token objective:
\begin{equation}
\mathcal{L}_{OPD}(\theta)
=
\mathbb{E}_{x \sim \mathcal{D},\, y \sim \pi_\theta(\cdot \mid x)}
\left[
\sum_{t=1}^{T}
-A_t \cdot \log \pi_\theta(y_t \mid s_t)
\right],
\label{eq:opd}
\end{equation}
where the advantage $A_t
=
\log \pi_T(y_t \mid s_t)
-
\log \pi_\theta(y_t \mid s_t)$ is treated as a constant in the gradient.

As discussed in Sec.~\ref{sec:intro}, recent work has identified several undesirable behaviors in OPD, such as repeated self-correction~\citep{luo2026demystifying,yang2026eos,fu2026revisiting}, ineffective supervision on later tokens~\citep{zhang2026fast,ziheng2026less}, and unreliable supervision once the student commits to an incorrect direction~\citep{xu2026pass,jiang2026trajectory}.
To study the necessity of on-policy sampling for distillation, we propose a simple alternative, Semi-OPD, in Sec.~\ref{sec:semi_opd_method}.

\begin{figure*}[t]
  \centering
  \includegraphics[width=0.95\linewidth]{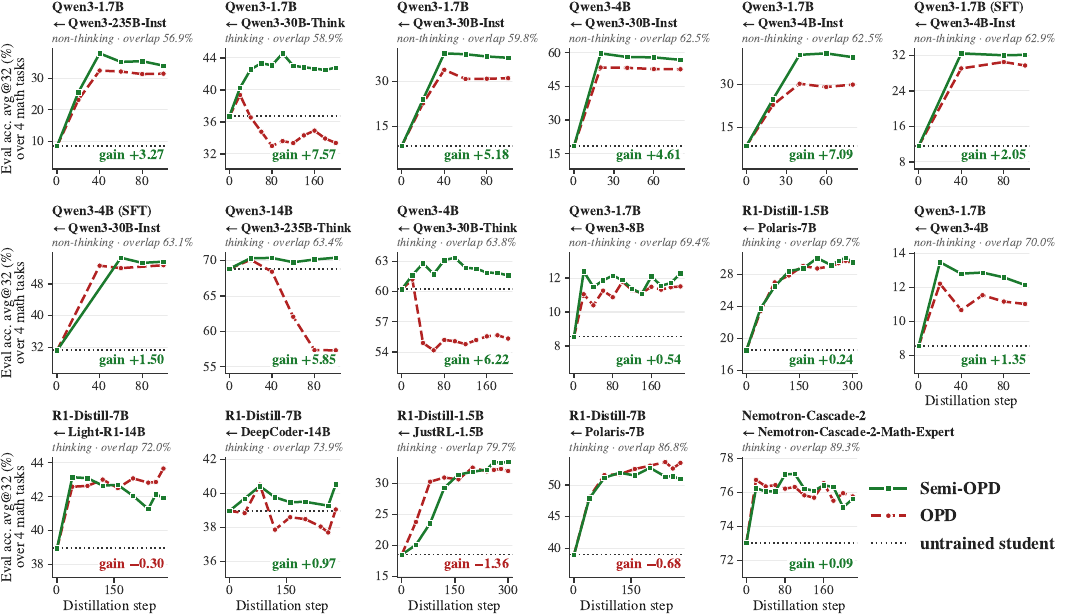}
\caption{
\textbf{\method vs.\ OPD across 17 model pairs.} Each panel represents one student--teacher pair, with its title containing the pair's initial top-64 overlap. Accuracy is measured as the mean avg@32 over AIME 2024/2025/2026 and HMMT February 2026. The number in each panel is the gain plotted in Fig.~\ref{fig:overlap}a, defined as the area between the two curves divided by the training horizon.
  }
  \label{fig:allpairs}
  \vspace{-1em}
\end{figure*}

\subsection{\method: Formulation}
\label{sec:semi_opd_method}

Let $\pi_0$ denote the initial student and $\pi_\theta$ the student being trained.
OPD samples each training response from the current student, $y \sim \pi_\theta(\cdot \mid x)$, so the training data changes as $\theta$ is
updated. \method instead samples responses once from the initial student before
training begins,
\begin{equation}
  \mathcal{D}_0 = \{(x_i, y_i)\}_{i=1}^{N}, \qquad y_i \sim \pi_0(\cdot \mid x_i).
\end{equation}
\method reuses this fixed set throughout training to minimize
the same objective as OPD:
\begin{equation}
  \mathcal{L}_{\mathrm{\method}}(\theta)
  = \mathbb{E}_{(x,y) \sim \mathcal{D}_0}
  \Big[\sum_{t=1}^{T} -A_t \cdot \log \pi_\theta(y_t \mid s_t)\Big].
  \label{eq:semi-opd}
\end{equation}
Although the rollouts are fixed, the student's log-probabilities, and hence the advantage $A_t$, are computed using the current parameters $\theta$ at each step, so the supervision signal continues to evolve.

\noindent \textbf{Training efficiency of \method.}
\method decouples data collection from optimization. We first generate all student rollouts from $\pi_0$; after these responses are generated, we run the teacher only once to compute the teacher log-probabilities; finally, we train the student on the fixed dataset with fixed teacher log-probabilities, without performing either online rollout generation or teacher forward passes. In addition, rollout generation, teacher scoring, and student optimization are executed as separate phases, each of which can use the full GPU cluster rather than partitioning resources among concurrent student generation, teacher serving, and training.

\noindent \textbf{Comparison with Lightning OPD.}
A prior work, Lightning OPD~\citep{wu2026lightning}, starts from a base student model and (1) performs supervised fine-tuning (SFT) on teacher rollouts and (2) performs distillation on offline student rollouts. Because it always starts from a base model, which cannot generate high-quality rollouts, it needs to retain step (1). Our \method essentially drops step (1) because we start from instruction-tuned student models, which aligns with real-world use cases. We show in Sec.~\ref{sec:experiments}/~\ref{sec:when_why} that \method outperforms Lightning OPD and that SFT on teacher rollouts has negative effects. In addition, we leverage our method to study when OPD should be used and the underlying reasons behind its failures, providing insights into best practices for OPD.

\vspace{-2mm}
\noindent \textbf{Top-$K$ overlap ratio for measuring initial teacher--student alignment.}
To measure the alignment between teacher--student pairs and cover diverse combinations in our experiments,
we follow \citep{li2026rethinking} and compare the student's and teacher's
top-$K$ token sets at each response position of the initial student's rollouts. Specifically, let   
$S_K(t) = \operatorname{TopK}(\pi_0(\cdot \mid s_t))$ and          
$T_K(t) = \operatorname{TopK}(\pi_T(\cdot \mid s_t))$ denote the student's   
and teacher's top-$K$ token sets, respectively. We define the top-$K$ overlap
ratio at position $t$ as
\[
\operatorname{Overlap@K}(t)
=
\frac{\left|S_K(t) \cap T_K(t)\right|}{K}.
\]
Because both sets contain $K$ tokens, this quantity is $1$ when the two
distributions have identical top-$K$ supports and $0$ when their supports are
disjoint. We report the token-weighted mean over all response tokens. Unless
otherwise noted, we use $K=64$.

\begin{table*}[t]
\centering
\setlength{\tabcolsep}{7pt}
\renewcommand{\arraystretch}{0.85}
\caption{
\method outperforms other OPD variants when the initial student--teacher overlap ratio is low, and remains comparable when the ratio is high.
Results are reported as avg@32 at temperature 1.0, averaged over the checkpoints at
steps 60/80/100. $^\S$Reported from~\citep{xu2026pass}.
}
\begin{tabular}{lccccc}
\toprule
Method
& AIME24 & AIME25 & AIME26 & HMMT Feb26 & Avg@32 \\
\midrule

\multicolumn{6}{l}{
\textbf{Qwen3-1.7B (Non-think) $\leftarrow$ Qwen3-4B-Instruct-2507 }
\hfill {\normalfont \underline{Initial overlap ratio: 62.5\%}}
} \\
\addlinespace[2pt]

\textit{Teacher}
& 60.4 & 46.0 & 52.2 & 31.2 & 47.5 \\
\textit{Student, untrained}
& 12.6 & 9.6 & 7.4 & 6.3 & 9.0 \\

\addlinespace[1pt]
Standard OPD$^\S$
& 35.8 & 25.5 & 23.3 & 20.1 & 26.2 \\
Speculative KD~\citep{xuspeculative}$^\S$
& 33.1 & 30.7 & 28.9 & 20.1 & 28.2 \\
FastOPD~\citep{zhang2026fast}$^\S$
& 42.3 & 30.4 & 26.4 & 23.6 & 30.7 \\
Lightning OPD~\citep{wu2026lightning}
& 41.0 & 28.8 & 27.5 & 22.7 & 30.0 \\
Relay-OPD~\citep{xu2026pass}$^\S$
& 42.7 & 32.8 & 30.5 & 24.7 & 32.7 \\
\textbf{\method (ours)}
& \textbf{50.3}
& \textbf{38.9}
& \textbf{41.8}
& \textbf{28.0}
& \textbf{39.8} \\

\midrule

\multicolumn{6}{l}{
\textbf{Qwen3-4B (Non-think) $\leftarrow$ Qwen3-30B-A3B-Instruct-2507}
\hfill {\normalfont \underline{Initial overlap ratio: 62.5\%}}
} \\
\addlinespace[2pt]

\textit{Teacher}
& 72.9 & 62.0 & 72.8 & 40.6 & 62.1 \\
\textit{Student, untrained}
& 23.6 & 18.0 & 16.8 & 14.9 & 18.3 \\

\addlinespace[1pt]
Standard OPD
& 55.1 & 45.3 & 49.1 & 30.0 & 44.9 \\
FastOPD~\citep{zhang2026fast}
& 63.5 & 55.5 & 56.2 & 34.9 & 52.5 \\
Lightning OPD~\citep{wu2026lightning}
& 64.0 & 56.8 & 54.3 & 33.5 & 52.2 \\
Relay-OPD~\citep{xu2026pass}
& 61.1 & 48.1 & 52.5 & 32.8 & 48.6 \\
\textbf{\method (ours)}
& \textbf{66.4}
& \textbf{63.1}
& \textbf{62.9}
& \textbf{37.4}
& \textbf{57.4} \\

\midrule





\multicolumn{6}{l}{
\textbf{NM-Cascade-2-30B-A3B $\leftarrow$ NM-Cascade-2-Math-Expert}
\hfill {\normalfont \underline{Initial overlap ratio: 89.3\%}}
} \\
\addlinespace[2pt]

\textit{Teacher}
& 82.8 & 80.0 & 76.4 & 58.2 & 74.4 \\
\textit{Student, untrained}
& 77.9 & 72.4 & 68.8 & 49.3 & 67.1 \\

\addlinespace[1pt]
Standard OPD
& 83.0 & 80.1 & 75.8 & 57.5 & 74.1 \\
FastOPD~\citep{zhang2026fast}
& 83.0 & \textbf{80.7} & \textbf{77.1} & \textbf{58.4} & \textbf{74.8} \\
Lightning OPD~\citep{wu2026lightning}
& \textbf{83.2} & 79.1 & 75.7 & 56.8 & 73.7 \\
Relay-OPD~\citep{xu2026pass}
& \textbf{83.2} & 80.1 & 73.3 & 57.7 & 73.6 \\
\textbf{\method (ours)}
& 83.0 & 79.7 & 77.0 & 56.8 & 74.1 \\

\bottomrule
\end{tabular}
\label{tab:baselines}
\vspace{-0.5em}
\end{table*}

\subsection{Benchmarking \method vs. OPD and its Variants}
\label{sec:experiments}
\vspace{-2mm}






\noindent \textbf{Experiment setup.}
We conduct experiments on 17 student--teacher pairs (full model-pair names are provided in Tab.~\ref{tab:allpairs} in the appendix), spanning the Qwen3~\citep{qwen3}, DeepSeek-R1-Distill~\citep{deepseekai2025deepseekr1incentivizingreasoningcapability}, and Nemotron-Cascade-2~\citep{yang2026nemotron} families. Qwen3 students (1.7B, 4B, and 14B) are distilled from Instruct and Thinking teachers with up to 235B-A22B parameters, using either original or SFT-warm-started initializations; DeepSeek-R1-Distill students (1.5B and 7B) are distilled from RL-trained descendants of the same checkpoints, including JustRL~\citep{he2025justrl}, Polaris~\citep{Polaris2025}, Light-R1~\citep{lightr1proj}, and DeepCoder~\citep{deepcoder2025}; and the Nemotron-Cascade-2 pair consists of the student checkpoint before OPD and its math-domain teacher used for the OPD step in~\citep{yang2026nemotron}. The pairs cover both non-thinking and thinking modes, with training and evaluation matched within each pair. All models are distilled on DAPO-Math~\citep{yu2025dapo}. Mathematical capability is measured by $\mathrm{avg}@32$ accuracy over AIME 2024--2026 and HMMT February 2026, using evaluation rollouts capped at 32k tokens (40k for Nemotron-Cascade-2), while cross-domain generalization is assessed on HumanEval and HumanEval+~\citep{chen2021evaluating}. More details are provided in Appendix~\ref{app:setup}.

\begin{figure*}[t]
  \centering
  \includegraphics[width=\linewidth]{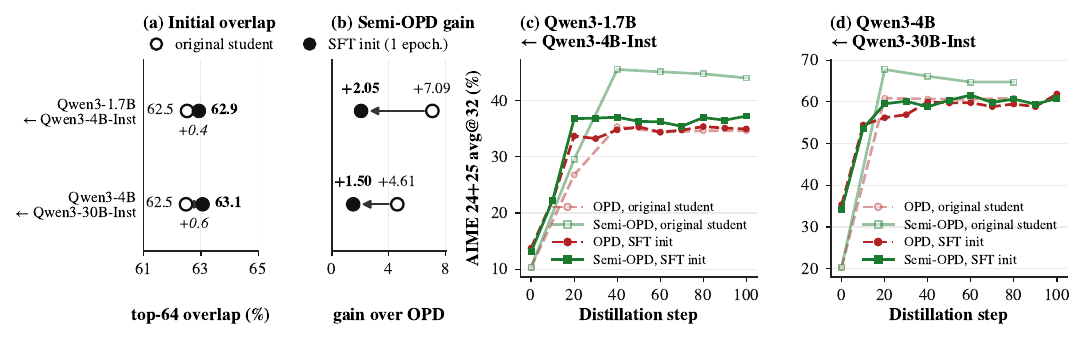}
\vspace{-1.5em}
\caption{
\textbf{Teacher-rollout SFT improves overlap ratios but reduces final performance after \method.} \textbf{(a)} SFT on teacher rollouts slightly increases the initial top-64 overlap. \textbf{(b)} SFT on teacher rollouts followed by \method still outperforms OPD but underperforms \method alone. \textbf{(c, d)} SFT on teacher rollouts negatively impacts final performance when combined with \method.
}

  \label{fig:sft}
  \vspace{-1em}
\end{figure*}

\textbf{\method vs.\ OPD across 17 model pairs.} Fig.~\ref{fig:allpairs} visualizes the average accuracy evolution during training achieved by \method and OPD across all 17 teacher--student pairs. We order the pairs by the overlap ratio  of the initial teacher--student pair, which is annotated above each plot. We also report the average accuracy gain of \method over OPD, defined as the area between the \method and OPD learning curves divided by their shared training horizon.

We observe that \method outperforms OPD on 14 of 17 pairs with improvements of up to $\sim$13\% over OPD at the last step, and the largest gains are generally achieved on pairs with lower initial overlap. On the remaining three pairs, \method still achieves comparable performance, with gaps of 0.1--2.6\%. The difference is especially pronounced for Qwen3 thinking-mode pairs: OPD undergoes a substantial performance collapse during training, whereas \method continues to improve.

\noindent \textbf{Comparison with OPD variants in accuracy and training efficiency.} In addition to standard OPD, we also compare \method with recent OPD variants across three model pairs with different model sizes, overlap ratios, and reasoning modes in Tab.~\ref{tab:baselines}.
We observe that (1) on the Qwen3-1.7B (non-thinking) and Qwen3-4B (non-thinking) students, which have low teacher--student overlap ratios, \method outperforms all other baselines in accuracy while also being the most efficient, e.g., achieving +7.1\% and +8.8\% higher accuracy with 35$\times$ and 19.5$\times$ less training time than Relay-OPD on the two pairs, respectively, according to Fig.~\ref{fig:speedup} and Tab.~\ref{tab:speedup}. (2) On the Nemotron-Cascade-2 pair, which has a high overlap ratio of approximately 89\%, our \method remains competitive with other OPD baselines and falls only slightly behind the best baseline by 0.7\% accuracy, while reducing the training time by 2.2x as compared to FastOPD and 4.5x compared to OPD.

The overall accuracy--training-efficiency trade-off is shown in Fig.~\ref{fig:speedup}, where \method achieves up to an 11.4$\times$ speedup over standard OPD and a 2.7$\times$ speedup over the fastest baseline, FastOPD. In contrast, another OPD variant, Relay-OPD, which also aims to make student rollouts closer to the teacher, requires interleaved teacher and student generation, resulting in lower efficiency.

\vspace{-0.5em}
\section{How to Choose Between OPD and \method?}
\vspace{-0.5em}
\label{sec:when_why}

As observed in Sec.~\ref{sec:experiments} and Fig.~\ref{fig:allpairs}, the relative performance of OPD and \method is highly correlated with the teacher--student overlap ratio. It can therefore serve as an indicative metric for determining when to use OPD or \method. We characterize this correlation more formally here.

\noindent \textbf{OPD is needed for teacher--student pairs with high overlap ratios; otherwise, \method is better and faster.} We compute the correlation between \method's gain over OPD, measured as the area between their learning curves normalized by their shared training horizon, and the overlap ratio of each teacher--student pair in Fig.~\ref{fig:overlap}. The two quantities are strongly correlated (Pearson's $r=-0.73$ across 17 pairs): pairs with lower initial overlap ratios tend to benefit substantially more from \method, while OPD is only needed for pairs with high overlap ratios.

This pattern implies that, when the student  differs substantially from the teacher, repeatedly sampling from the evolving student policy can cause online rollouts to degrade (several examples are shown in Appendix~\ref{app:qualitative}) and weaken the supervision signal; reusing rollouts from the initial student can provide a more stable distillation signal.  We dive into the underlying reason in Sec.~\ref{sec:why}.

\textbf{Key insights.}
The above phenomenon provides useful implications for the practical use and research of OPD: (1) Given a target teacher--student pair and target-domain prompts, determine whether to use OPD or \method based on their overlap ratio, rather than blindly using OPD; (2) For future studies of improved OPD schemes in LM post-training pipelines, carefully select the teacher--student pair used as the experimental setting; otherwise, \method can serve as a strong baseline.

\textbf{Does SFT on teacher rollouts close the gap?}
We also experiment with two pairs in which the students undergo one epoch of SFT on teacher-generated rollouts, followed by OPD or \method. This matches the setup of Lightning OPD~\citep{wu2026lightning}. As shown in Fig.~\ref{fig:sft}, we observe that (1) SFT on teacher rollouts slightly increases teacher--student overlap ratios, while \method still outperforms OPD on both pairs, albeit by a smaller margin; and (2) Our \method without the additional teacher-rollout SFT stage can outperform \method with this stage, indicating that SFT on teacher rollouts is unnecessary.
We also note that, while~\citep{wu2026lightning} frames fixed rollouts primarily as an efficient approximation to OPD, our results show that freezing the rollout cache can improve distillation when the initial teacher--student overlap is low.

\begin{figure*}[t]
  \centering
  \includegraphics[width=\linewidth]{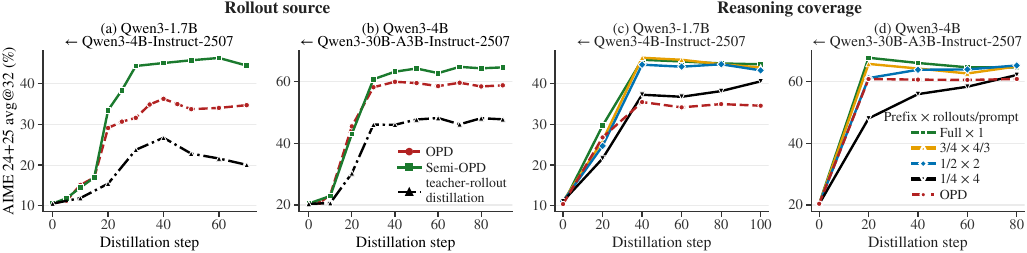}
  \vspace{-1em}
  \caption{
\textbf{Both rollout source and reasoning coverage matter for distillation.}
\textbf{(a, b)} \method outperforms teacher-rollout distillation (and OPD) on both pairs. \textbf{(c, d)} We compare \method trained on the first $1/4$, $1/2$, $3/4$, or all of each rollout, while keeping the total number of training tokens fixed by scaling the number of rollouts per prompt. Distillation on all stages performs best.
}
  \label{fig:source_coverage}
  \label{fig:reasoning_coverage}
  \vspace{-0.5em}
\end{figure*}

\begin{figure}[t]
  \centering
  \includegraphics[width=\linewidth]{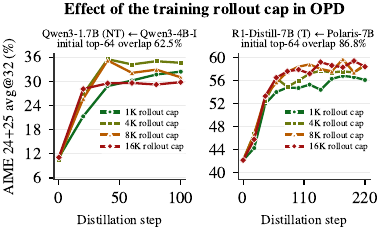}
  \vspace{-1.5em}
  \caption{
  Longer rollouts hurt the low-overlap pair (left) but not the high-overlap pair (right).
  }
  \vspace{-1em}
  \label{fig:rollout_cap}
\end{figure}

\begin{figure*}[t]
  \centering
  \vspace{-0.5em}
  \begin{minipage}[t]{0.48\linewidth}
    \centering
    \includegraphics[width=\linewidth]{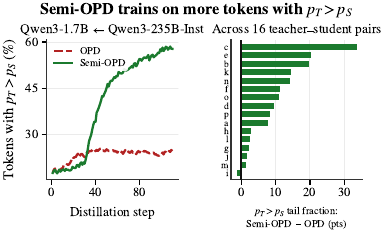}
    \captionsetup{skip=4pt}
    \caption{\textbf{The percentage of tokens with positive rewards}, i.e., fraction of training
    tokens with $p_T > p_S$ under \method and OPD.}
    \label{fig:teacher_student_signal}
  \end{minipage}\hfill
  \begin{minipage}[t]{0.48\linewidth}
    \centering
    \includegraphics[width=\linewidth]{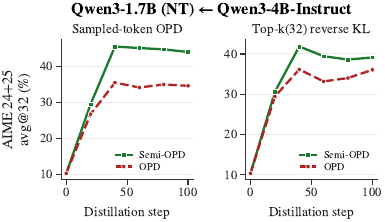}
    \captionsetup{skip=4pt}
    \caption{\textbf{\method vs.\ OPD under two distillation objectives}, i.e.,
    the sampled-token and top-$k$ reverse KL objectives.}
    \label{fig:distillation_objective}
  \end{minipage}
\end{figure*}

\begin{figure*}[t]
  \centering
  \includegraphics[width=\linewidth]{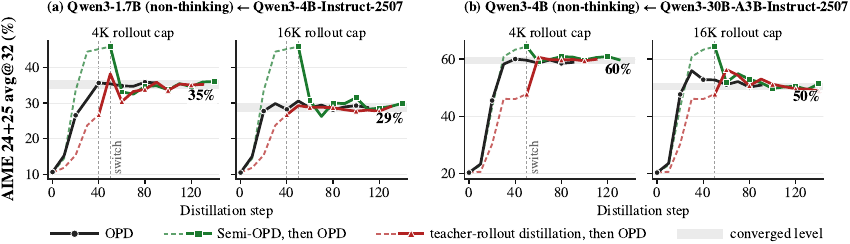}
    \vspace{-1.5em}
\caption{
\textbf{Switching to OPD converges toward a rollout-determined equilibrium, largely independent of initialization.}
OPD eval accuracy trajectories when initialized from the initial student, a \method checkpoint, or a teacher-rollout distillation checkpoint. Switching to OPD can either decrease or improve performance (see \cref{fig:cache_switch_highoverlap}), with trajectories approaching the performance level of OPD trained from scratch under the same rollout cap. Dashed lines indicate training before switching to OPD.
}
  \label{fig:handoff}
  \vspace{-1em}
\end{figure*}

\vspace{-0.5em}
\section{Why Does OPD Often Underperform \method?}
\vspace{-0.5em}
\label{sec:why}

We further investigate why OPD underperforms \method in low-overlap cases.

\noindent \textbf{Our hypothesis.}
We hypothesize that effective distillation requires training data (i.e., rollouts) to satisfy two properties: (1) on-policyness w.r.t. the student to avoid exposure bias, and (2) high-quality supervision from the teacher, which is often determined by on-policyness w.r.t. the teacher distribution, i.e., how well the rollouts align with the teacher's own inference-time distribution.
OPD prioritizes the former while struggling with the latter, especially for low-overlap teacher--student pairs; \method trades off some student on-policyness for more effective supervision signals, as analyzed below.

More specifically, for low-overlap pairs, rollouts become increasingly longer during OPD and exhibit various artifacts~\citep{luo2026demystifying, yang2026eos, fu2026revisiting}. This makes later tokens in a rollout increasingly off-policy w.r.t. the teacher's distribution, making teacher supervision on them unhelpful or even harmful~\citep{xu2026pass,jiang2026trajectory}.
Truncating the rollouts and training only on the prefix can mitigate this issue~\citep{zhang2026fast,ziheng2026less}, but at the cost of removing supervision on later tokens. As demonstrated later, covering supervision signals across all reasoning stages in full trajectories is important for final performance.
In contrast, \method trains on rollouts from the initial student, which are usually shorter than on-policy rollouts, thereby better maintaining on-policyness w.r.t. the teacher while still covering full trajectories with all reasoning stages.
We support the above analysis through the following study.

\noindent \textbf{Student on-policyness matters: \method vs.\ teacher-rollout distillation.}
We study the importance of student on-policyness to \method by replacing the initial-student rollout cache with teacher-generated rollouts while keeping all other settings unchanged.
As shown in Fig.~\ref{fig:source_coverage}(a, b), \method consistently outperforms teacher-rollout distillation. This indicates that the benefit of \method does not come merely from using a fixed rollout cache. Rather, retaining student-generated trajectories to maintain student on-policyness is also important.

\noindent \textbf{Training on longer rollouts can hurt OPD for low-overlap pairs.}
As shown in \cref{fig:rollout_cap}, we vary the rollout length (i.e., the rollout cap) used for OPD training and observe that increasing the rollout cap can hurt the low-overlap pair while benefiting the higher-overlap pair.
We hypothesize that, for low-overlap pairs, training on long rollouts that are off-policy w.r.t. the teacher may tend to reinforce superficial reasoning patterns without sufficiently improving the underlying reasoning process, causing subsequent student rollouts to deteriorate (examples comparing rollouts are in Appendix~\ref{app:qualitative}).

\noindent \textbf{Supervision signals covering all reasoning stages are essential.}
If long rollouts can be harmful, one simple solution is to truncate them, at the cost of dropping supervision from later reasoning stages.
To isolate the value of reasoning-stage coverage, we perform an ablation by truncating the cached \method traces to the first $1/4$, $1/2$, or $3/4$ of their length, while scaling the number of rollouts per prompt to keep the total training token budget fixed. As shown in Fig.~\ref{fig:reasoning_coverage}(c, d), restricting supervision to the first quarter substantially slows convergence and reduces final accuracy, whereas training on full trajectories achieves the best peak accuracy. Thus, both covering all reasoning stages and maintaining on-policyness w.r.t. the teacher are important.

\noindent \textbf{\method exposes the student to more tokens preferred by the teacher.}
\method exactly satisfies the above requirement by providing full trajectory coverage with shorter rollout lengths. To analyze the resulting distillation signal, we measure the percentage of training tokens with $p_T > p_S$, i.e., tokens that receive a positive reward due to higher probabilities assigned by the teacher.
As shown in Fig.~\ref{fig:teacher_student_signal}, \method maintains a larger fraction of such tokens than OPD throughout training. This indicates that \method exposes the student to more teacher-preferred tokens, i.e., more on-policy w.r.t. the teacher, whereas OPD more often produces signals that suppress the student's own sampled tokens.
Importantly, \method also outperforms OPD under top-$k$ reverse KL, as shown in Fig.~\ref{fig:distillation_objective}, indicating that its advantage is not specific to the sampled-token objective.

Together, these findings suggest that \method benefits from preserving student-anchored rollouts that remain effective teacher supervision and broad coverage of the reasoning process. These advantages can outweigh the benefits of fully on-policy sampling under large teacher--student mismatch.

\vspace{-0.5em}
\section{More Intriguing Observations} 
\label{sec:training_dynamics}
\vspace{-0.5em}

Beyond the main comparison, we observe several interesting phenomena that provide a more complete picture of \method versus OPD. More interesting observations are provided in the appendix.

\begin{figure}[h]
  \centering
  \includegraphics[width=0.8\linewidth]{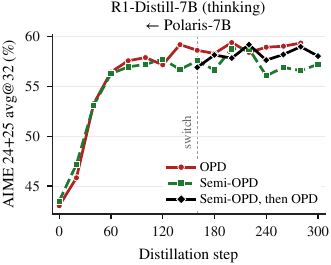}
  \vspace{-0.5em}
  \caption{Switching from \method to OPD on a high-overlap pair with an initial top-64 overlap ratio of 86.8\% closes the gap with OPD applied from scratch.}
  \label{fig:cache_switch_highoverlap}
\end{figure}

\noindent \textbf{OPD converges to a rollout-determined equilibrium, regardless of initialization.}
A natural question is whether OPD can further improve after \method.
We study this by initializing OPD from \method or teacher-rollout distillation checkpoints as warm starts. As shown in Fig.~\ref{fig:handoff}, on low-overlap pairs, regardless of initialization, all runs eventually converge to the performance level reached by OPD initialized from the initial student model under the same rollout cap, while the warm start only affects the convergence trajectory. 
In contrast, for a high-overlap pair, as shown in Fig.~\ref{fig:cache_switch_highoverlap}, switching from \method to OPD improves performance and closes most of the gap with OPD applied from scratch.
This suggests that repeated on-policy sampling drives distillation toward an equilibrium determined by the online rollout data distribution rather than the initial checkpoint. Characterizing this equilibrium remains an interesting direction for future work.

\begin{figure}[h]
  \centering
  \includegraphics[width=\linewidth]{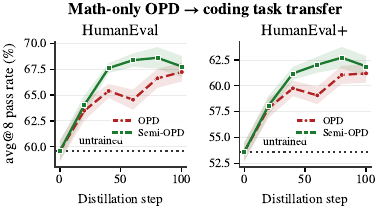}
  \vspace{-1.5em}
  \caption{\textbf{Transfer to coding.} Code performance of models
    distilled on math only (Qwen3-1.7B $\leftarrow$ Qwen3-4B-Instruct-2507).}
  \label{fig:code_transfer}
  \vspace{-1em}
\end{figure}

\noindent \textbf{The gains transfer beyond the training domain.}
As shown in Fig.~\ref{fig:code_transfer}, models distilled only on math-heavy datasets~\citep{yu2025dapo} also improve on HumanEval and HumanEval+, with \method remaining ahead of OPD.
This suggests that the gains are not limited to fitting the math training distribution but transfer to broader reasoning tasks.


\begin{figure}[h]
  \centering
  \includegraphics[width=\linewidth]{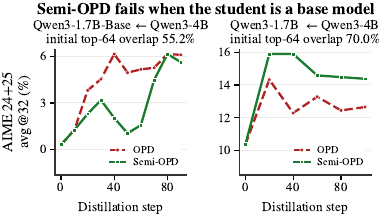}
  \vspace{-1em}
   \caption{\textbf{Initial student competence.} \method vs.\ OPD with the same
    teacher, from base (left) and instruction-tuned (right) students.}
    \label{fig:base_student}
    \vspace{-1em}
\end{figure}

\noindent \textbf{\method requires adequately strong initial rollouts.}
Since \method trains only on rollouts from $\pi_0$, the initial rollouts need to meet a minimum quality threshold for effective distillation. For example, as shown in Fig.~\ref{fig:base_student}, with a base student, the initial rollouts are too weak and \method underperforms OPD; with an instruction-tuned student, \method regains its advantage.
Thus, \method is most suitable when the initial student already produces meaningful trajectories.
\vspace{-2mm}
\section{Related work}
\vspace{-2mm}
\label{sec:related}

\textbf{On-Policy Distillation and Its Mechanics.}
OPD trains a student on trajectories sampled from its own policy, while a teacher provides dense supervision through reverse-KL or related objectives~\citep{agarwal2024policy,gu2024minillm,lu2025onpolicydistillation, song2026survey}.
By optimizing on the student's own visitation distribution, OPD mitigates the exposure bias of offline distillation~\citep{hinton2015distillingknowledgeneuralnetwork, kim2016sequence}, an idea shared with interactive imitation learning methods such as DAgger~\citep{ross2011reduction}.
OPD has recently become a popular post-training stage for reasoning models~\citep{xiao2026mimov2flashtechnicalreport,qwen3,xu2026deepseek,team2026kimi}, motivating recent work to better understand its underlying mechanics~\citep{jin2026entropy, fu2026revisiting, armandpour2026unmasking}.
For example, successful OPD is linked to teacher--student compatibility~\citep{li2026rethinking}, while even a single training example or 1\% of tokens can recover much of full-data OPD~\citep{fu2026rethinkingii, sheng20261}.

\textbf{Failure Modes in OPD.}
Recent work has identified several failure modes of OPD: student rollouts can grow longer, fall into repeated self-correction, and fail to terminate~\citep{luo2026demystifying,yang2026eos,fu2026revisiting,yu2026mismatch}, and later portions of a rollout can drift away from the teacher distribution, yielding increasingly unreliable supervision~\citep{ziheng2026less,xu2026pass,jiang2026trajectory,wang2026demystifying}.
Existing methods mitigate these issues by truncating student rollouts~\citep{zhang2026fast,ziheng2026less}, letting the teacher intervene during generation~\citep{xuspeculative,xu2026pass}, refining student trajectories with the teacher~\citep{jiang2026trajectory}, warm-starting OPD on teacher data~\citep{li2026rethinking}, or constraining updates to a trust region~\citep{xing2026trust,xie2026trust}.
Other work relaxes strict on-policyness  for efficiency, using fixed rollouts from a teacher-SFT reference policy~\citep{wu2026lightning,wu2026lightningii} or policy-lagged student rollouts in asynchronous training~\citep{kang2026asyncopd}.
We study the simplest form of this relaxation: rollouts are sampled once from the initial student, without any teacher-SFT stage, and no refreshing.
Beyond efficiency gain, we find that freezing rollouts can also improve accuracy.
Moreover, rather than focusing on a few model pairs, we evaluate 17 diverse model pairs and show that this benefit is strongly correlated with the initial teacher--student top-$k$ overlap, with lower-overlap pairs benefiting more.

\vspace{-2mm}

\section{Conclusion}
\vspace{-2mm}

This work revisits when on-policy sampling is truly necessary for LM distillation. By systematically comparing OPD with the proposed \method across diverse teacher--student pairs, we show that continually regenerating student rollouts is often unnecessary and can even be harmful, especially when the teacher and student are poorly aligned. Our analyses highlight a key principle: effective distillation requires balancing on-policyness w.r.t. both the student and the teacher, with the initial teacher--student overlap serving as a practical indicator of when to use OPD or \method. Beyond providing an efficient alternative to OPD, our findings motivate future work to rethink on-policy sampling as a design choice rather than a default, and to develop distillation methods that explicitly account for teacher--student alignment and rollout quality.

{
  \small
  \bibliographystyle{unsrt}
  \bibliography{ref}
}

\onecolumn
\appendix
\section{Training and Evaluation Setup}

\label{app:setup}

\paragraph{Models.}
Tab.~\ref{tab:allpairs} lists all 17 teacher--student pairs used in our experiments. They cover three groups:
(i)~Qwen3 students (1.7B, 4B, 14B) to be distilled from Qwen3 Instruct-2507 and Thinking-2507 teachers of up to 235B-A22B parameters;
(ii)~DeepSeek-R1-Distill-Qwen students (1.5B, 7B) to be distilled from RL-trained models of the same checkpoints (JustRL, Polaris, Light-R1, DeepCoder);
(iii)~one Nemotron Cascade-2 math expert pair.
Pairs (g) and (h) start from an SFT checkpoint of the student, after doing 1 epoch of SFT on the teacher generated rollouts.
Reasoning mode (thinking T / non-thinking NT) is fixed and matched per pair. Both methods of a pair train and evaluate with the same chat template and thinking mode, and the evaluation mode always matches the training mode.
The DeepSeek-R1-Distill chat templates always prefill \texttt{<think>}, so we treat those pairs as thinking-mode.

\paragraph{Training data and setup.}
All runs train on the prompts from the
\href{https://huggingface.co/datasets/open-r1/DAPO-Math-17k-Processed}
{DAPO-Math-17K dataset}
(14{,}116 English prompts after deduplication), using 128 prompts per step
and one rollout per prompt.
One epoch is therefore around 110 steps.
For \method, we sample one response per prompt from the initial student $\pi_0$ before training, using the same sampling parameters and response cap as OPD.
OPD reads the prompts in the same order.
As a result, the two methods see the same prompts at every step and differ only in which model generated the response.
Both methods use the same sampled-token reverse-KL objective.
They also share the teacher, learning rate, batch size and evaluation protocol.
Tab.~\ref{tab:hparams} lists the hyperparameters. Additionally, it is worth noting that in our work, Fast OPD is implemented with a training rollout limit of 4K, following~\citep{xu2026pass}. For only the non-thinking pairs, we also adopt a 4K training rollout length for OPD in the comparisons shown in \cref{fig:allpairs} and \cref{fig:overlap}. In \cref{tab:baselines}, however, OPD is trained with a 16K rollout limit for all pairs to clearly distinguish with Fast OPD. For exact rollout shape used during training can refer to \cref{tab:allpairs} too. Note that this is for training rollout, for evaluation we all use 32k or 40k limit.

\begin{table}[h]
  \centering
  \small
  \caption{Training and evaluation hyperparameters. }
  \label{tab:hparams}
  \begin{tabular}{lcc}
    \toprule
    & Non-thinking pairs & Think/Long-CoT pairs \\
    \midrule
    \multicolumn{3}{l}{\emph{Optimization}} \\
    Optimizer & \multicolumn{2}{c}{AdamW ($\beta_1{=}0.9$, $\beta_2{=}0.999$, wd $0.01$)} \\
    Learning rate & \multicolumn{2}{c}{$1\times10^{-6}$, constant (no warm-up)} \\
    Gradient clipping & \multicolumn{2}{c}{1.0} \\
    Precision & \multicolumn{2}{c}{bfloat16} \\
    Prompts per step / rollouts per prompt & \multicolumn{2}{c}{128 / 1} \\
    Global batch size & \multicolumn{2}{c}{128} \\
    \midrule
    \multicolumn{3}{l}{\emph{Rollouts}} \\
    Max prompt length & \multicolumn{2}{c}{2{,}048} \\
    Max response length (training) & 4{,}096 & 16{,}384 \\
    Sampling temperature / top-$p$ & \multicolumn{2}{c}{1.0 / 1.0} \\
    \midrule
    \multicolumn{3}{l}{\emph{Loss}} \\
    Objective & \multicolumn{2}{c}{Sampled-token reverse KL} \\
    \midrule
    \multicolumn{3}{l}{\emph{Evaluation}} \\
    Benchmarks & \multicolumn{2}{c}{AIME24, AIME25, AIME26, HMMT Feb26} \\
    Metric & \multicolumn{2}{c}{avg@32} \\
    Temperature / top-$p$ & \multicolumn{2}{c}{1.0 / 1.0} \\
    Max response length & \multicolumn{2}{c}{32k (40k for Nemotron model)} \\
    \bottomrule
  \end{tabular}
\end{table}

\paragraph{Infrastructure.}
We train with NeMo-RL (DTensor backend) and generate rollouts and evaluation samples with vLLM.
Runs use 1--8 nodes of 8$\times$\,H100 GPUs, depending on the teacher size.

\providecommand{\hf}[2]{\href{https://huggingface.co/#1}{#2}}
\begin{table}[h]
  \centering
  \small
  \caption{All 17 teacher--student pairs, sorted by AUC gain of \method over OPD.
  Model names link to the exact Hugging Face checkpoints used.
  Mode: T = thinking / long chain-of-thought, NT = non-thinking.
  Cap: maximum response length of training rollouts, shared by both \method and OPD.
  Overlap: top-64 overlap between the initial student and the teacher, in \%.
  Last-step: \method minus OPD at the last step both methods evaluated.
  AUC: signed area between the two accuracy curves over the steps both methods evaluated, divided by the length of that span.
  Gains are avg@32 points on the four-benchmark mean.
  R1-Distill = DeepSeek-R1-Distill-Qwen; Cascade-2 = Nemotron-Cascade-2
  (SFT): the linked student after one epoch of SFT on teacher rollouts.
  $^\dagger$Evaluated with a 40{,}960-token response limit; all other pairs use 32{,}768.}
  \label{tab:allpairs}
 \resizebox{\ifdim\width>\linewidth\linewidth\else\width\fi}{!}{%
  \begin{tabular}{cllccrrr}
    \toprule
    & & & & Train Len & Init. & \multicolumn{2}{c}{Gain} \\
    \cmidrule(lr){7-8}
    \# & Student & Teacher & Mode & Cap & Overlap & Last-step & AUC \\
    \midrule
    n & \hf{Qwen/Qwen3-1.7B}{Qwen3-1.7B} & \hf{Qwen/Qwen3-30B-A3B-Thinking-2507}{Qwen3-30B-A3B-Thinking} & T & 16K & 58.9 & \textbf{+9.4} & \textbf{+7.57} \\
    a & \hf{Qwen/Qwen3-1.7B}{Qwen3-1.7B} & \hf{Qwen/Qwen3-4B-Instruct-2507}{Qwen3-4B-Instruct-2507} & NT & 4K & 62.5 & \textbf{+9.5} & \textbf{+7.09} \\
    p & \hf{Qwen/Qwen3-4B}{Qwen3-4B} & \hf{Qwen/Qwen3-30B-A3B-Thinking-2507}{Qwen3-30B-A3B-Thinking} & T & 16K & 63.8 & \textbf{+6.2} & \textbf{+6.22} \\
    o & \hf{Qwen/Qwen3-14B}{Qwen3-14B} & \hf{Qwen/Qwen3-235B-A22B-Thinking-2507}{Qwen3-235B-A22B-Thinking} & T & 16K & 63.4 & \textbf{+13.1} & \textbf{+5.85} \\
    b & \hf{Qwen/Qwen3-1.7B}{Qwen3-1.7B} & \hf{Qwen/Qwen3-30B-A3B-Instruct-2507}{Qwen3-30B-A3B-Instruct-2507} & NT & 4K & 59.8 & \textbf{+6.8} & \textbf{+5.18} \\
    f & \hf{Qwen/Qwen3-4B}{Qwen3-4B} & \hf{Qwen/Qwen3-30B-A3B-Instruct-2507}{Qwen3-30B-A3B-Instruct-2507} & NT & 4K & 62.5 & \textbf{+4.2} & \textbf{+4.61} \\
    c & \hf{Qwen/Qwen3-1.7B}{Qwen3-1.7B} & \hf{Qwen/Qwen3-235B-A22B-Instruct-2507}{Qwen3-235B-A22B-Instruct-2507} & NT & 4K & 56.9 & \textbf{+2.6} & \textbf{+3.27} \\
    g & \hf{Qwen/Qwen3-1.7B}{Qwen3-1.7B} (SFT) & \hf{Qwen/Qwen3-4B-Instruct-2507}{Qwen3-4B-Instruct-2507} & NT & 4K & 62.9 & \textbf{+2.4} & \textbf{+2.05} \\
    h & \hf{Qwen/Qwen3-4B}{Qwen3-4B} (SFT) & \hf{Qwen/Qwen3-30B-A3B-Instruct-2507}{Qwen3-30B-A3B-Instruct-2507} & NT & 4K & 63.1 & +0.9 & +1.50 \\
    d & \hf{Qwen/Qwen3-1.7B}{Qwen3-1.7B} & \hf{Qwen/Qwen3-4B}{Qwen3-4B} & NT & 4K & 70.0 & +1.1 & +1.35 \\
    m & \hf{deepseek-ai/DeepSeek-R1-Distill-Qwen-7B}{R1-Distill-7B} & \hf{agentica-org/DeepCoder-14B-Preview}{DeepCoder-14B} & T & 16K & 73.9 & +1.5 & +0.97 \\
    e & \hf{Qwen/Qwen3-1.7B}{Qwen3-1.7B} & \hf{Qwen/Qwen3-8B}{Qwen3-8B} & NT & 4K & 69.4 & +0.8 & +0.54 \\
    j & \hf{deepseek-ai/DeepSeek-R1-Distill-Qwen-1.5B}{R1-Distill-1.5B} & \hf{POLARIS-Project/Polaris-7B-Preview}{Polaris-7B} & T & 16K & 69.7 & +0.1 & +0.24 \\
    q & Cascade-2-30B-A3B & Cascade-2-Math-Expert$^{\dagger}$ & T & 16K & 89.3 & $-$0.1 & +0.09 \\
    l & \hf{deepseek-ai/DeepSeek-R1-Distill-Qwen-7B}{R1-Distill-7B} & \hf{qihoo360/Light-R1-14B-DS}{Light-R1-14B} & T & 16K & 72.0 & $-$1.7 & $-$0.30 \\
    k & \hf{deepseek-ai/DeepSeek-R1-Distill-Qwen-7B}{R1-Distill-7B} & \hf{POLARIS-Project/Polaris-7B-Preview}{Polaris-7B} & T & 16K & 86.8 & $-$2.6 & $-$0.68 \\
    i & \hf{deepseek-ai/DeepSeek-R1-Distill-Qwen-1.5B}{R1-Distill-1.5B} & \hf{hbx/JustRL-DeepSeek-1.5B}{JustRL-1.5B} & T & 16K & 79.7 & +1.4 & $-$1.36 \\
    \bottomrule
  \end{tabular}}
\end{table}

\clearpage
\newpage

\section{Training Dynamics Across All Pairs}
\label{app:allpairs_dynamics}

\begin{figure}[H]
  \centering
  \includegraphics[width=0.85\linewidth]{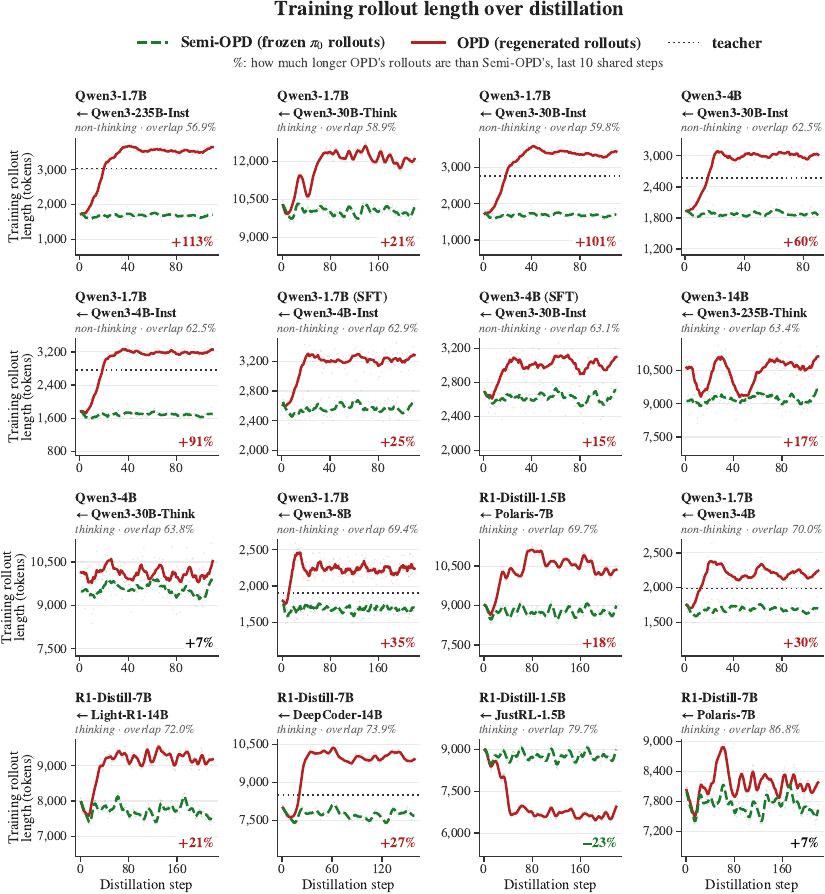}
  \caption{\textbf{Training rollout length.} Mean generated tokens per
  training rollout. The number in each panel is how much longer OPD's rollouts
  are than \method's over the last 10 shared steps; the dotted line is the
  teacher's length, where available.}
  \label{fig:allpairs_length}
\end{figure}

\begin{figure}[H]
  \centering
  \includegraphics[width=0.85\linewidth]{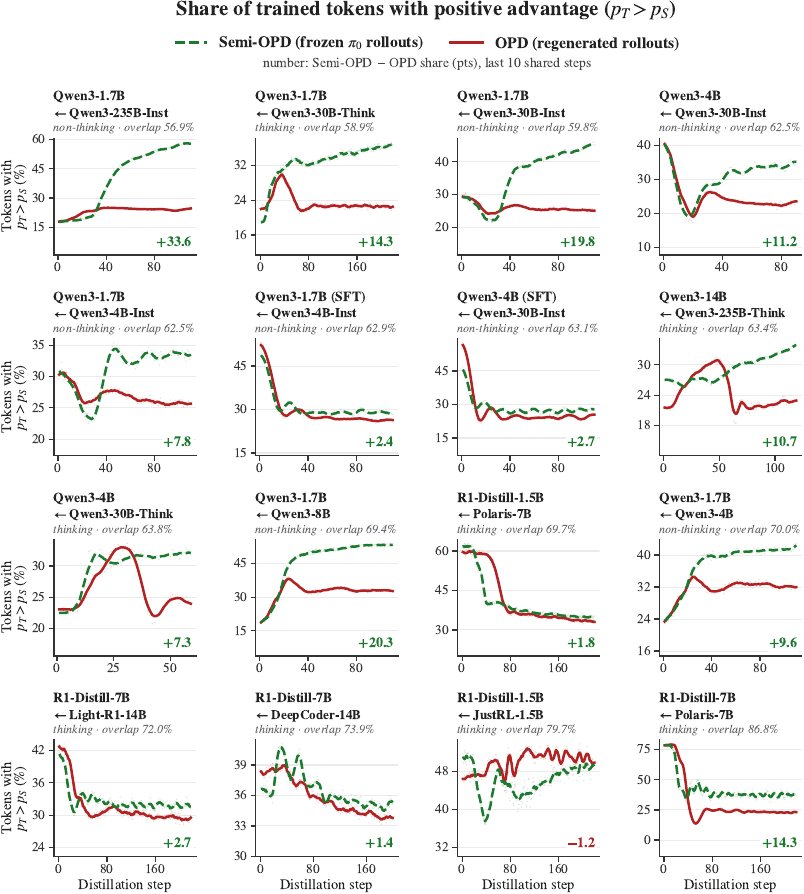}
  \caption{\textbf{Positively supervised tokens.} Share of training tokens
  with $p_T > p_S$. The number in each panel is \method minus OPD, in points,
  over the last 10 shared steps.}
  \label{fig:allpairs_signal}
\end{figure}

\begin{figure}[H]
  \centering
  \includegraphics[width=0.85\linewidth]{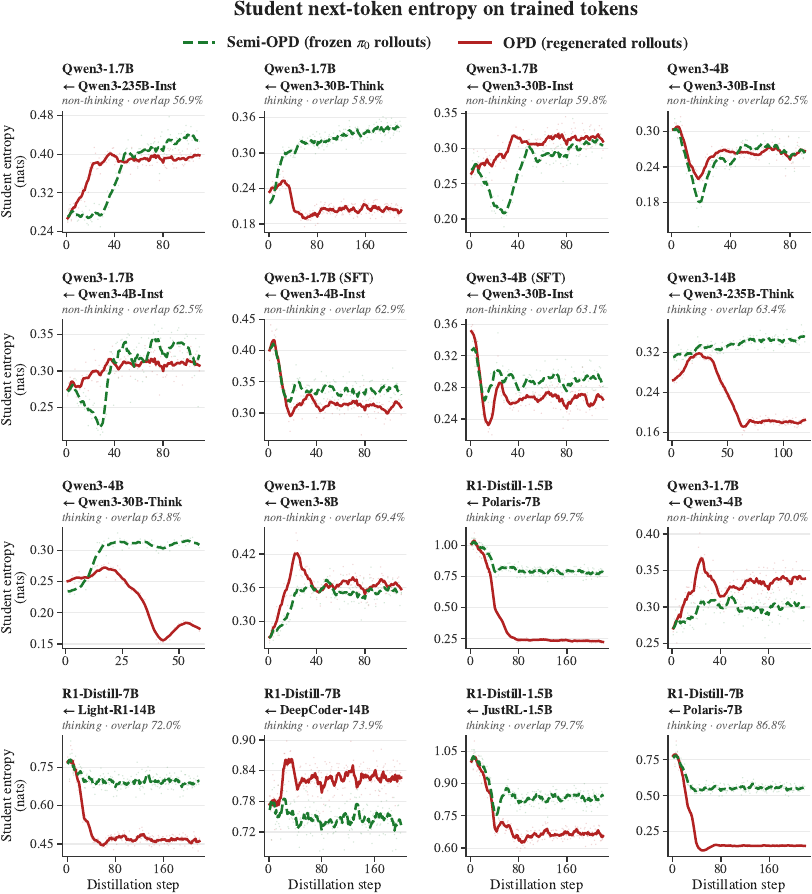}
  \caption{\textbf{Student entropy.} Mean next-token entropy of the student
  (nats) on the training tokens.}
  \label{fig:allpairs_student_entropy}
\end{figure}

\begin{figure}[H]
  \centering
  \includegraphics[width=0.85\linewidth]{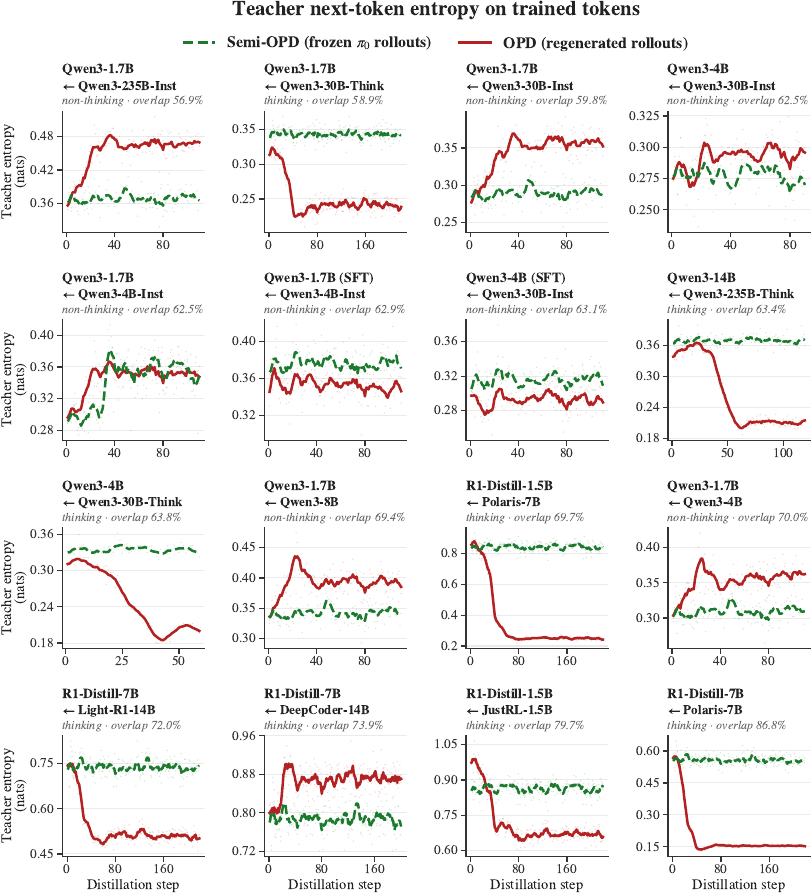}
  \caption{\textbf{Teacher entropy.} Mean next-token entropy of the teacher
  (nats) on the same training tokens, i.e.\ how uncertain the teacher is on
  the rollouts it supervises.}
  \label{fig:allpairs_teacher_entropy}
\end{figure}

\begin{figure}[H]
  \centering
  \includegraphics[width=0.85\linewidth]{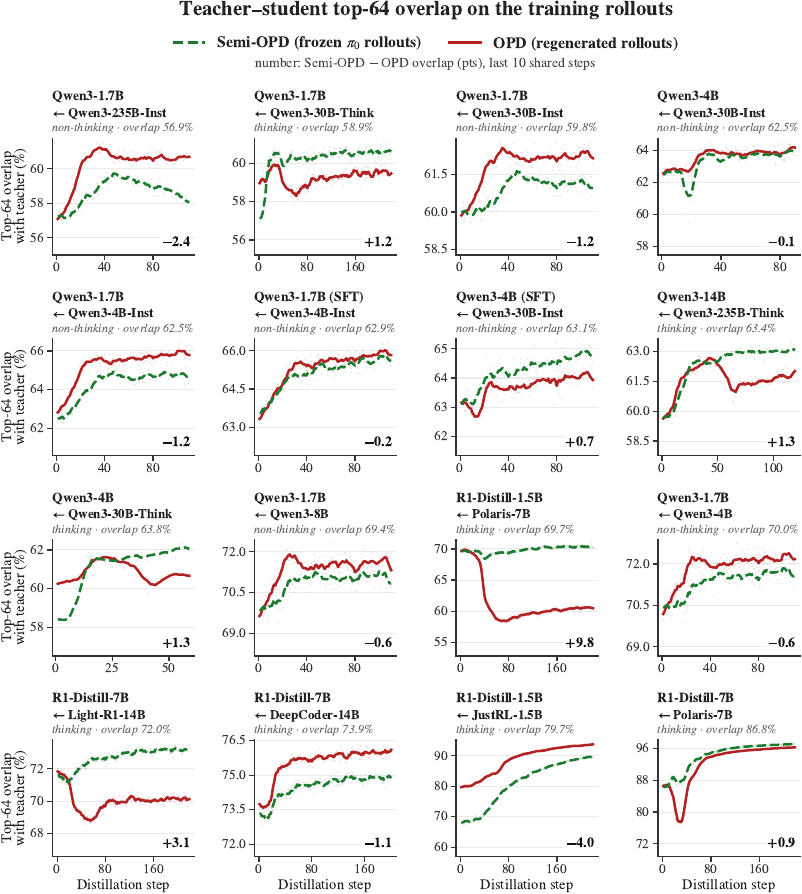}
  \caption{\textbf{Teacher--student overlap on the training rollouts.} Top-64 overlap
  between the current student and the teacher, computed on each method's own training
  tokens with the same metric as the initial overlap in \cref{sec:method}. The number in
  each panel is \method minus OPD, in points, over the last 10 shared steps. For the two
  SFT-initialized pairs, overlap was measured only at initialization and at step 110.}
  \label{fig:allpairs_overlap}
\end{figure}

We provide more training dynamics discussion in this section. \Cref{fig:allpairs_length,fig:allpairs_signal,fig:allpairs_student_entropy,fig:allpairs_teacher_entropy,fig:allpairs_overlap}
track five properties of the training rollouts for 16 of the 17 pairs (all except the
Nemotron-Cascade-2 pair).

\paragraph{Rollout length.}
As shown in \cref{fig:allpairs_length}, OPD's rollouts lengthen as training proceeds,
while \method's stay at the length of the initial student's rollouts by construction.
Over the last 10 shared steps, OPD's rollouts are 7--113\% longer than \method's in 15 of
16 pairs. The growth is largest for the low-overlap non-thinking Qwen3 pairs, where OPD's
rollouts often exceed the teacher's own length. The only exception is
R1-Distill-1.5B $\leftarrow$ JustRL-1.5B, whose OPD rollouts get shorter.

\newpage
\paragraph{Positively supervised tokens.}
As shown in \cref{fig:allpairs_signal}, \method keeps a larger share of positively
supervised tokens ($p_T > p_S$) than OPD in 15 of 16 pairs. The exception is again
R1-Distill-1.5B $\leftarrow$ JustRL-1.5B.

\clearpage
\newpage
\paragraph{Student and teacher entropy.}
As shown in \cref{fig:allpairs_student_entropy,fig:allpairs_teacher_entropy}, on thinking
pairs OPD drives both the student's and the teacher's next-token entropy on the training
tokens well below \method's in 7 of 8 pairs (the exception is
R1-Distill-7B $\leftarrow$ DeepCoder-14B), i.e., OPD increasingly trains on tokens that
both models already predict with high confidence. On non-thinking pairs, student entropy
stays similar between the two methods, while teacher entropy is usually higher under OPD,
indicating that OPD's rollouts drift toward regions where the teacher is less certain.

\newpage
\paragraph{Teacher--student overlap.}
As shown in \cref{fig:allpairs_overlap}, the top-64 overlap between the current student
and the teacher on the training tokens generally rises under both methods as the student moves
toward the teacher, but the ordering differs across different model pairs.

\newpage

\section{Training Cost Comparison}
Semi-OPD is faster because rollout generation and teacher scoring are done once before training. Similar to Lightning OPD~\citep{wu2026lightning}, this removes online rollout generation and teacher inference from the training loop, so all GPUs can be used for rollout generation, teacher log-probability scoring, and student training. For a fixed student, the same rollout cache can also be reused across different teachers; evaluating a new teacher only requires computing its log-probabilities on the cached rollouts. This can further reduce the cost of comparing or selecting teachers in practice. In contrast, Relay-OPD requires interleaved student and teacher generation throughout training, leading to higher training cost.

\begin{table*}[h]
  \centering
  \caption{\textbf{Training cost.} H100 GPU-hours for 100 training steps, excluding evaluation.
  Semi-OPD's cost is the sum of rollout generation, teacher scoring, and student training.
  FastOPD is OPD with rollouts truncated to 4K tokens. Speedup is the method's GPU-hours
  divided by Semi-OPD's.}
  \vspace{2mm}
  \label{tab:speedup}
  \setlength{\tabcolsep}{10pt}
  \renewcommand{\arraystretch}{1.15}
  \begin{tabular}{@{}lrr@{}}
    \toprule
    Method & GPU-hours & Semi-OPD speedup \\
    \midrule
    \multicolumn{3}{@{}l}{\textit{Qwen3-1.7B (non-thinking) $\leftarrow$ Qwen3-4B-Instruct-2507}$^{\dagger}$} \\
    \quad Relay-OPD & 173.3 & $34.7\times$ \\
    \quad OPD       & 56.8  & $11.4\times$ \\
    \quad FastOPD   & 11.7  & $2.3\times$ \\
    \quad \textbf{Semi-OPD} & \textbf{5.0} & -- \\
    \midrule
    \multicolumn{3}{@{}l}{\textit{Qwen3-4B (non-thinking) $\leftarrow$ Qwen3-30B-A3B-Instruct-2507}$^{\dagger}$} \\
    \quad Relay-OPD & 604.4 & $19.5\times$ \\
    \quad OPD       & 199.1 & $6.4\times$ \\
    \quad FastOPD   & 77.3  & $2.5\times$ \\
    \quad \textbf{Semi-OPD} & \textbf{31.0} & -- \\
    \midrule
    \multicolumn{3}{@{}l}{\textit{Qwen3-14B (thinking) $\leftarrow$ Qwen3-235B-A22B-Thinking-2507}} \\
    \quad Relay-OPD & 1546.5 & $24.6\times$ \\
    \quad OPD       & 480.0  & $7.6\times$ \\
    \quad FastOPD   & 169.8  & $2.7\times$ \\
    \quad \textbf{Semi-OPD} & \textbf{63.0} & -- \\
    \midrule
    \multicolumn{3}{@{}l}{\textit{Nemotron-Cascade-2-30B-A3B $\leftarrow$ Nemotron-Cascade-2-Math-Expert}} \\
    \quad Relay-OPD & 933.5 & $14.1\times$ \\
    \quad OPD       & 301.4 & $4.5\times$ \\
    \quad FastOPD   & 147.5 & $2.2\times$ \\
    \quad \textbf{Semi-OPD} & \textbf{66.4} & -- \\
    \bottomrule
  \end{tabular}\\[4pt]
  \begin{minipage}{0.85\linewidth}
  \footnotesize
  $^{\dagger}$Semi-OPD uses 4K training rollouts on these pairs and OPD uses 16K; on the other
  two pairs both use 16K. Training rollout caps are listed in \cref{tab:allpairs}.\\
  \end{minipage}
\end{table*}

\clearpage
\newpage
\section{Length and Self-Correction Inflation}
\label{app:inflation}

\begin{figure*}[h]
    \centering
    \includegraphics[width=\linewidth]{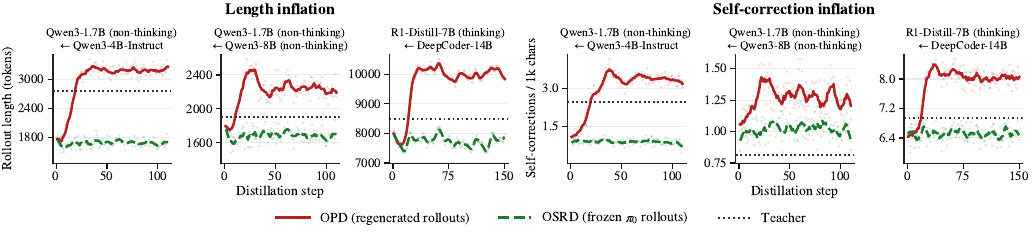}
    \caption{\textbf{On-policy rollouts inflate during OPD.}
    Rollout length (left three panels) and self-correction word density (right
    three panels) of the training rollouts. OPD rollouts grow longer and use more
    self-correction words, eventually exceeding the teacher (dotted), while the
    frozen rollouts sampled once from the initial student $\pi_0$ do not.}
    \label{fig:inflation}
\end{figure*}

Fig.~\ref{fig:inflation} tracks two properties of the rollouts each method
trains on: rollout length (mean generated tokens per rollout) and
self-correction word density. Under OPD, the rollouts are regenerated by the
current student at every step. Under Semi-OPD, they are frozen rollouts
sampled  from the initial student $\pi_0$.

\paragraph{Measuring self-correction.}
We measure self-correction with a lexical word density. We count
case-insensitive, whole-word occurrences of a fixed set of self-correction
words, \{\textit{wait, but, hmm, actually, hold, however, yet, oh,
alternatively, no, ah, oops, well}\}, in the generated completion (prompt
excluded). At each training step, we pool all 128 training rollouts and report
$1000 \times \sum_i h_i / \sum_i c_i$, where $h_i$ is the number of word hits
and $c_i$ the number of characters in rollout~$i$. The teacher reference
(dotted line) uses the same definition on 128 teacher samples ($T{=}1.0$, same
rollout cap) on the final training step's prompts; rollout length for the
teacher is measured on the same samples. Because the vocabulary includes common
discourse words such as \textit{but} and \textit{no}, the metric is a proxy for
reflective, hedging style rather than a count of verified corrections. We show more qualitative examples of rollout comparisons between OPD and Semi-OPD in Appendix~\ref{app:qualitative}.

\clearpage
\newpage
\section{Qualitative Examples: OPD versus Semi-OPD Training Rollouts}
\label{app:qualitative}

\providecommand{\refl}[1]{\textcolor{red}{\textbf{#1}}}
\providecommand{\capped}{{\color{gray}\textit{[generation cap reached; no final answer]}}}
\providecommand{\gap}[1]{{\color{gray}\textit{[\,\dots\ #1 tokens omitted\,\dots]}}}
\newtcolorbox{qualproblem}{colback=gray!6,colframe=gray!50,boxrule=0.4pt,arc=1pt,
  left=3pt,right=3pt,top=2pt,bottom=2pt,fontupper=\footnotesize}
\newtcolorbox{Semi-OPDbox}[1]{colback=white,colframe=green!45!black,boxrule=0.5pt,arc=1pt,
  left=2pt,right=2pt,top=1pt,bottom=1pt,toptitle=1pt,bottomtitle=1pt,
  fonttitle=\scriptsize\bfseries,fontupper=\scriptsize,title={#1},
  before upper={\setlength{\parskip}{2pt}\setlength{\parindent}{0pt}}}
\newtcolorbox{opdbox}[1]{colback=white,colframe=black!65,boxrule=0.5pt,arc=1pt,
  left=2pt,right=2pt,top=1pt,bottom=1pt,toptitle=1pt,bottomtitle=1pt,
  fonttitle=\scriptsize\bfseries,fontupper=\scriptsize,title={#1},
  before upper={\setlength{\parskip}{2pt}\setlength{\parindent}{0pt}}}

\begin{figure*}[h]
\centering
\begin{qualproblem}
\textbf{Example 1.} Qwen3-1.7B $\leftarrow$ Qwen3-30B-A3B-Thinking (thinking). OPD rollout at training step 201; Semi-OPD rollout from $\pi_0$; generation cap 16{,}384 tokens.\par\smallskip
\textbf{Problem.} What is the hundreds digit of the sum below?

$1 + 12 + 123 + 1234 + 12345 + 123456 + 1234567 + 12345678 + 123456789$
\end{qualproblem}
\vspace{-1pt}
\begin{minipage}[t]{0.495\linewidth}\vspace{0pt}
\begin{Semi-OPDbox}{Semi-OPD (frozen $\pi_0$ rollout): 5{,}688 tokens, 36 reflection words, reaches an answer}
\gap{1{,}101}

Now, take mod 1000: 205.

\refl{Wait}, so the total sum mod 1000 is 205? So the last three digits of the sum are 205. Therefore, the hundreds digit is 2. Is that correct?

\gap{2{,}535}

So total sum is 137,174,205.

Now, the last three digits are 205, so the hundreds digit is 2. So that confirms it.

\gap{1{,}380}

Therefore, the hundreds digit of the sum is 2.

\textbf{Final Answer} \newline\rule{0pt}{2.6ex}$\boxed{2}$

\gap{550}
\end{Semi-OPDbox}
\end{minipage}\hfill
\begin{minipage}[t]{0.495\linewidth}\vspace{0pt}
\begin{opdbox}{OPD (step 201): 16{,}384 tokens, 80 reflection words, hits the cap}
\gap{6{,}594}

Sum of hundreds digits: 0+0+1+2+3+4+5+6+7 = 28

Add carryover from tens: 28 + 4 = 32

Hundreds digit of sum: 32 mod 10 = 2

\gap{2{,}658}

Now, the hundreds place of the sum is the second digit from the right, so 32 mod 100 is 32, so the hundreds digit is 3

\gap{2{,}458}

\refl{Ah}! Here's the mistake: when I said the sum of the hundreds digits is 28, plus carryover from tens (4), that's the total sum for the hundreds place, which is 32. \refl{But} the hundreds place of the sum is the digit we get from 32, which is the tens digit of 32, so 2, and the carryover is 3 (from 32 to the thousands place).

\gap{3{,}097}

\refl{Wait}, no. 32 is the sum of the digits in the hundreds place (plus carryover), which is the value of the hundreds place in the total sum. For instance, if the hundreds place sum is 32, it means that the total sum has a 3 in the hundreds place and 2 in the thousands place? \refl{No}, the place values are:

- If you have 32 as the sum for the hundreds place, it means that you have 3*100 + 2*1, so the hundreds digit is 3, and the tens digit is 2.

\gap{1{,}140}

For example, if the sum of the hundreds digits plus carryover is 32, then the sum is 32*100 + (remainder), so the hundreds digit is 2 (the 32's tens place).

\refl{Wait}, no: if the sum of the hundreds digits plus carryover is 32, then the total value contributed by the hundreds place is 32*100. \refl{But} in reality, when you add numbers, the hundreds place is the 100s place, so the hundreds digit is the digit in the 100s place of the

\capped
\end{opdbox}
\end{minipage}

\label{fig:qual_t1}
\end{figure*}

\begin{figure*}[p]
\centering
\begin{qualproblem}
\textbf{Example 2.} DeepSeek-R1-Distill-Qwen-7B $\leftarrow$ DeepCoder-14B (thinking). OPD rollout at training step 128; Semi-OPD rollout from $\pi_0$; generation cap 16{,}384 tokens.\par\smallskip
\textbf{Problem.} Find the positive solution to \newline $\frac 1{x^2-10x-29}+\frac1{x^2-10x-45}-\frac 2{x^2-10x-69}=0.$
\end{qualproblem}
\vspace{-1pt}
\begin{minipage}[t]{0.495\linewidth}\vspace{0pt}
\begin{Semi-OPDbox}{Semi-OPD (frozen $\pi_0$ rollout): 2{,}228 tokens, 10 reflection words, reaches an answer}
\gap{176}

Let me let $y = x^2 - 10x$. Then, each denominator becomes $y - 29$, $y - 45$, and $y - 69$. So, substituting that in, the equation becomes:

$\frac{1}{y - 29} + \frac{1}{y - 45} - \frac{2}{y - 69} = 0.$

\gap{769}

So, the equation reduces to:

-64y +2496=0.

\gap{718}

\refl{But} the question asks for positive solution, so x=13.

\gap{438}
\end{Semi-OPDbox}
\end{minipage}\hfill
\begin{minipage}[t]{0.495\linewidth}\vspace{0pt}
\begin{opdbox}{OPD (step 128): 9{,}016 tokens, 117 reflection words, reaches an answer}
\gap{124}

Let's set $y = x^2 - 10x$. That seems like a smart substitution because each denominator can be written in terms of y.

\gap{1{,}916}

\refl{Wait}, perhaps it's a perfect square? Let me check if 257,664 is a square.

\refl{Well}, 508$^2$=258,064.

507$^2$=257,049.

Subtract 257,664 -257,049=615.

Not a perfect square. \refl{Hmm}\dots{} \refl{Wait}, perhaps 615 can be written as... Not sure.

\gap{735}

\refl{Wait}, perhaps the initial approach is getting me stuck because the square root is messy. Maybe I made a mistake in earlier calculation? Let me go back.

\refl{Wait}, \refl{hold on}. Let me recap:

\gap{139}

\refl{Wait}, \refl{hold on}, perhaps I made a mistake in expanding the denominator?

\refl{Wait}, (y -29)(y -45)= y$^2$ -45y -29y +29$\times$45= y$^2$ -74y +1305?

\refl{Wait}, 29$\times$45: 20$\times$45=900, 9$\times$45=405, total 1305, that's correct.

\refl{Wait}, could I make a mistake in cross-multiplying earlier? Let me double-check that.

\gap{5{,}172}

Therefore, okay, the logic is hold.

So steps: substituted y=x$^2$ -10x, made the equation only in y, solved for y=39, found x=13 and x=-3, recognized which is required: positive.

Hence, the positive solution is 13.

\gap{587}
\end{opdbox}
\end{minipage}

\label{fig:qual_ds7}
\end{figure*}

\begin{figure*}[p]
\centering
\begin{qualproblem}
\textbf{Example 3.} Qwen3-1.7B $\leftarrow$ Qwen3-30B-A3B-Instruct-2507 (non-thinking). OPD rollout at training step 90; Semi-OPD rollout from $\pi_0$; generation cap 4{,}096 tokens.\par\smallskip
\textbf{Problem.} Let $f(x) = \frac{x+1}{x-1}$. Then for $x^2 \neq 1$, find the expression for $f(-x)$. The original answer is in $\frac{k}{m}$ format, please give the value of $k + m$.
\end{qualproblem}
\vspace{-1pt}
\begin{minipage}[t]{0.495\linewidth}\vspace{0pt}
\begin{Semi-OPDbox}{Semi-OPD (frozen $\pi_0$ rollout): 308 tokens, 0 reflection words, reaches an answer}
\gap{69}

Step 1: Substitute $-x$ into $f(x)$

$f(-x) = \frac{(-x) + 1}{(-x) - 1}$

Simplify numerator and denominator:

$f(-x) = \frac{-x + 1}{-x - 1}$

We can factor out a negative sign from both the numerator and denominator:

$f(-x) = \frac{-(x - 1)}{-(x + 1)} = \frac{x - 1}{x + 1}$

Step 2: Identify $k$ and $m$

We have:

$f(-x) = \frac{x - 1}{x + 1}$

This is already in the form $\frac{k}{m}$, where:

$k = -1,\allowbreak \quad m = 1$

Step 3: Compute $k + m$

$k + m = -1 + 1 = \boxed{0}$

Final Answer:

$\boxed{0}$
\end{Semi-OPDbox}
\end{minipage}\hfill
\begin{minipage}[t]{0.495\linewidth}\vspace{0pt}
\begin{opdbox}{OPD (step 90): 4{,}096 tokens, 87 reflection words, hits the cap}
\gap{239}

Step 3: Check if it's in the form $\frac{k}{m}$

This is not a constant, it's a rational function in $x$, but we are to assume the form is $\frac{k}{m}$, which is a constant. \refl{Wait} --- perhaps we misread?

\gap{156}

\refl{Wait} --- but $f(-x)$ is not a constant function. Only when $x$ is a constant --- but $x$ is a variable.

Unless --- maybe the question is a trick?

\refl{Wait} --- let's recheck.

\gap{151}

\refl{Wait} --- maybe the question is not asking for $f(-x)$, but rather the expression for $f(-x)$ in a simplified form that becomes a constant?

\refl{Wait} --- but $f(-x) = \frac{x - 1}{x + 1}$ is not constant.

\gap{510}

\refl{Wait} --- \textbf{maybe the original question has a typo?}

\gap{531}

\refl{Wait} --- press on.

\refl{Wait} --- \textbf{let's suppose that the }original answer is $\frac{1}{2}$

\gap{466}

\refl{Wait} --- \textbf{finally!} Let's suppose that the question is misworded.

\gap{1{,}658}

\refl{Wait} --- perhaps it was supposed to be the \textbf{value} of $f(-x)$ at a specific $x$? \refl{But} no.

\refl{Wait} --- maybe the question is asking for the expression of $f(-x)$ in terms of $f(x)$?

\gap{100}

\capped
\end{opdbox}
\end{minipage}

\label{fig:qual_q30i}
\end{figure*}

\FloatBarrier

\end{document}